\documentclass[10pt,twocolumn,letterpaper]{article}

\usepackage{cvpr}
\usepackage{times}
\usepackage{epsfig}
\usepackage{graphicx}
\usepackage{amsmath, mathtools}
\usepackage{amssymb}

\usepackage{multirow}
\usepackage{makecell}
\usepackage{color}
\usepackage[dvipsnames]{xcolor}
\usepackage[font=normal]{subfig}
\usepackage[export]{adjustbox}
\usepackage{setspace}
\usepackage{booktabs}
\usepackage{array}
\usepackage{arydshln}   

\usepackage[pagebackref=true,breaklinks=true,letterpaper=true,colorlinks,bookmarks=false]{hyperref}

\cvprfinalcopy 

\def\cvprPaperID{47} 

\ifcvprfinal\fi

\begin{document}

\title{NTIRE 2020 Challenge on Image and Video Deblurring}
\author{
    Seungjun Nah \and Sanghyun Son \and Radu Timofte \and Kyoung Mu Lee
    \and Yu Tseng \and Yu-Syuan Xu \and Cheng-Ming Chiang \and Yi-Min Tsai      
    \and Stephan Brehm \and Sebastian Scherer                                   
    \and Dejia Xu \and Yihao Chu \and Qingyan Sun                               
    \and Jiaqin Jiang \and Lunhao Duan \and Jian Yao                            
    \and Kuldeep Purohit \and Maitreya Suin \and A.N. Rajagopalan               
    \and Yuichi Ito                                                             
    \and Hrishikesh P S \and Densen Puthussery \and Akhil K A \and Jiji C V     
    \and Guisik Kim                                                             
    \and Deepa P L 
    \and Zhiwei Xiong \and Jie Huang \and Dong Liu                              
    \and Sangmin Kim \and Hyungjoon Nam \and Jisu Kim \and Jechang Jeong        
    \and Shihua Huang                                                           
    \and Yuchen Fan \and Jiahui Yu \and Haichao Yu \and Thomas S. Huang         
    \and Ya Zhou \and Xin Li \and Sen Liu \and Zhibo Chen                       
    \and Saikat Dutta \and Sourya Dipta Das                                     
    \and Shivam Garg                                                            
    \and Daniel Sprague \and Bhrij Patel \and Thomas Huck                       
}

\maketitle
\thispagestyle{empty}

\begin{abstract}
    
    Motion blur is one of the most common degradation artifacts in dynamic scene photography.
    This paper reviews the NTIRE 2020 Challenge on Image and Video Deblurring.
    In this challenge, we present the evaluation results from 3 competition tracks as well as the proposed solutions.
    Track 1 aims to develop single-image deblurring methods focusing on restoration quality.
    On Track 2, the image deblurring methods are executed on a mobile platform to find the balance of the running speed and the restoration accuracy.
    Track 3 targets developing video deblurring methods that exploit the temporal relation between input frames.
    In each competition, there were 163, 135, and 102 registered participants and in the final testing phase, 9, 4, and 7 teams competed.
    The winning methods demonstrate the state-of-the-art performance on image and video deblurring tasks.
    
\end{abstract}

\section{Introduction}
\label{sec:intro}

    {\let\thefootnote\relax\footnotetext{S. Nah (seungjun.nah@gmail.com, Seoul National University), S. Son, R. Timofte, K. M. Lee are the NTIRE 2020 challenge organizers, while the other authors participated in the challenge. 
    \\Appendix~\ref{sec:appendix} contains the authors' teams and affiliations.
    \\NTIRE 2020 webpage:\\~\url{https://data.vision.ee.ethz.ch/cvl/ntire20/}}}

    As smartphones are becoming the most popular type of cameras in the world, snapshot photographs are prevailing. 
    Due to the dynamic nature of the hand-held devices and free-moving subjects, motion blurs are commonly witnessed on images and videos.
    Computer vision literature has studied post-processing methods to remove blurs from photography by parametrizing camera motion~\cite{gupta2010single, hirsch2011fast, whyte2012non} or generic motion~\cite{harmeling2010space, Ji_2012_CVPR, levin2007blind, Kim_2013_ICCV, Kim_2014_CVPR, Kim_2015_CVPR}.
    
    Modern machine learning based computer vision methods employ large-scale datasets to train their models. 
    For image and video deblurring tasks, 
    GOPRO~\cite{Nah_2017_CVPR}, DVD~\cite{Su_2017_CVPR}, WILD~\cite{noroozi2017motion} datasets were proposed as ways to synthesize blurry images by mimicking camera imaging pipeline from high-frame-rate videos. Recent dynamic scene deblurring methods train on such datasets to develop image~\cite{Nah_2017_CVPR, Tao_2018_CVPR, Zhang_2018_CVPR, Jin_2018_CVPR} and video~\cite{Wieschollek_2017_ICCV, Kim_2017_ICCV, Kim_2018_ECCV, Nah_2019_CVPR, Jin_2019_CVPR, chen2018reblur2deblur} deblurring methods.
    
    However, the early datasets for deblurring lacked in the quality of reference images and were limited in the realism of blur.
    With the NTIRE 2019 workshop, a new improved dataset, REDS~\cite{Nah_2019_CVPR_Workshops_REDS} was proposed by providing longer sensor read-out time, using a measured camera-response function and interpolating frames, \etc.
    REDS dataset was employed for NTIRE 2019 Video Deblurring~\cite{Nah_2019_CVPR_Workshops_Deblur} and Super-Resolution~\cite{Nah_2019_CVPR_Workshops_SR} Challenges. 

    Succeeding the previous year, NTIRE 2020 Challenge on Image and Video Deblurring presents 3 competition tracks.
    In track 1, single image deblurring methods are submitted and competed on desktop environments, focusing on the image restoration accuracy in terms of PSNR and SSIM. 
    In track 2, similarly to track 1, single image deblurring methods are submitted but deployed on a mobile device. Considering the practical application environment, the image restoration quality as well as the running time is evaluated together. 
    Track 3 exploits temporal information, continuing the NTIRE 2019 Video Deblurring Challenge.
    
    This challenge is one of the NTIRE 2020 associated challenges: deblurring, nonhomogeneous dehazing~\cite{ancuti2020ntire}, perceptual extreme super-resolution~\cite{zhang2020ntire}, video quality mapping~\cite{fuoli2020ntire}, real image denoising~\cite{abdelhamed2020ntire}, real-world super-resolution~\cite{lugmayr2020ntire}, spectral reconstruction from RGB image~\cite{arad2020ntire} and demoireing~\cite{yuan2020demoireing}. 
    
\section{Related Works}
\label{sec:related}
    The REDS dataset is designed for non-uniform blind deblurring. As all the methods submitted to the 3 tracks use deep neural networks, we describe the related works. Also, we describe the previous studies on neural network deployment on mobile devices.
    
    
    \subsection{Single Image Deblurring}
    \label{sec:related_image_deblur}
    Traditional energy optimization methods~\cite{harmeling2010space, Ji_2012_CVPR, levin2007blind, Kim_2013_ICCV, Kim_2014_CVPR} jointly estimated the blur kernel and the latent sharp image from a blurry image. 
    Early deep learning methods tried to obtain the blur kernel from neural networksby estimating local motion field~\cite{Sun_2015_CVPR, Gong_2017_CVPR} for later latent image estimation. 
    From the advent of the image deblurring datasets~\cite{Nah_2017_CVPR, Su_2017_CVPR}, end-to-end learning methods were presented by estimating the deblurred outputs directly~\cite{Nah_2017_CVPR,Tao_2018_CVPR,Zhang_2018_CVPR} without kernel estimation.
    From the initial coarse-to-fine architecture~\cite{Nah_2017_CVPR} whose parameters were scale-specific, scale-recurrent networks~\cite{Tao_2018_CVPR} and selective sharing scheme~\cite{Gao_2019_CVPR} has been proposed.
    Spatially varying operation was studied by using spatial RNN in \cite{Zhang_2018_CVPR}
    While the previous methods behave in class-agnostic manner, as face and human bodies are often being the main subject of photographs, class-specific deblurring methods were proposed~\cite{Shen_2019_ICCV, Ren_2019_ICCV}. They employ semantic information or 3D facial priors to better reconstruct target class objects.
    On another flow of studies, several approaches~\cite{Jin_2018_CVPR, Purohit_2019_CVPR} were proposed to extract multiple sharp images from a single blurry image. They are motivated by the fact that a blurry images have long exposure that could be an accumulation of many sharp frames with short exposure.
    
    
    
    
    
    
    
    
    
    
    
    \subsection{Video Deblurring}
    \label{sec:related_video_deblur}
    Video Deblurring methods exploit the temporal relation between the input frames in various manners.
    DBN~\cite{Su_2017_CVPR} stacks 5 consecutive frames in channels and a CNN learns to aggregate the information between the frames.
    The frames are aligned by optical flow before being fed into the network.
    Recurrent neural networks are used to pass the information from the past frames to the next ones~\cite{Wieschollek_2017_ICCV, Kim_2017_ICCV, Kim_2018_ECCV, Nah_2019_CVPR}.
    RDN~\cite{Wieschollek_2017_ICCV} and OVD~\cite{Kim_2017_ICCV} performs additional connections to the hidden state steps to better propagate useful information. STTN~\cite{Kim_2018_ECCV} was proposed on top of DBN and OVD to exploit long-range spatial correspondence information. 
    IFI-RNN~\cite{Nah_2019_CVPR} investigates into reusing parameters to make hidden states more useful.
    Recently proposed STFAN~\cite{Zhou_2019_ICCV} proposes filter adaptive convolution(FAC) to apply element-wise convolution kernels. As the convolutional kernel is acquired from the feature value, the operation is spatially varying.

    At the NTIRE 2019 Video Deblurring Challenge~\cite{Nah_2019_CVPR_Workshops_Deblur}, EDVR~\cite{Wang_2019_CVPR_Workshops} was the winning solution. PCD alignment module is devised to handle large motion and TSA fusion module is proposed to utilize the spatio-temporal attention.
    
    
    
    
    
    
    
    
    \subsection{Neural Network for Mobile Deployment}
    \label{sec:related_mobile}
    While deep neural network methods have brought significant success in computer vision, most of them require heavy computation and large memory footprint.
    For practical usage such as mobile deployment, lightweight and efficient model architectures are required~\cite{Ignatov_2018_ECCV_Workshops_AI,Ignatov_2019_ICCV_Workshops_AI}.
    Therefore, several methods have been developed to compress pre-trained neural networks while keeping their performances.
    
    For example, network quantization aims to reduce the number of bits for representing weight parameters and intermediate features.
    With a degree of performance loss, binary~\cite{courbariaux2016binarized, Rastegari_2016_ECCV} or ternary~\cite{li2016ternary} quantization significantly reduce model complexity and memory usage.
    To utilize efficient bit-shift operation, network parameters can be represented as powers of two~\cite{zhou2017incremental}.
    %

    On the other hand, several studies tried to limit the number of parameters by pruning~\cite{han2015deep, han2015learning, He_2017_ICCV} or using a set of parameters with indexed representations~\cite{chen2015compressing}. 
    Recent pruning methods consider convolution operation structure~\cite{li2016pruning, Luo_2017_ICCV, Son_2018_ECCV}.
    Also, as convolutional filters are multi-dimensional, the weights were approximated by decomposition techniques~\cite{jaderberg2014speeding, Son_2018_ECCV, zhang2015accelerating, Wang_2017_ICCVW}.
    
    On another point of view, several methods are aiming to design efficient network architectures.
    MobileNet V1~\cite{howard2017mobilenets}, V2~\cite{Sandler_2018_CVPR}, ThunderNet~\cite{Qin_2019_ICCV} adopt efficient layer compositions to build lightweight models in carefully handcrafted manners.
    Furthermore, MnasNet~\cite{Tan_2019_CVPR}, MobileNet V3~\cite{Howard_2019_ICCV}, EfficientNet~\cite{tan2019efficientnet} employ neural architecture search to find low-latency models while keeping high accuracy.
    Despite the success in architecture efficiency, most of the networks compression methods have been mostly focusing on high-level tasks such as image recognition and object detection.
    
    As general image restoration requires 8-bit depth at minimum, it is a nontrivial issue to apply the previous compression techniques to design lightweight models.
    Thus, \cite{Ma_2019_CVPR_Workshops} utilizes a local quantization technique in image super-resolution task by binarizing layers in the residual blocks only.
    Besides, there were attempts to apply efficient designs of MobileNets~\cite{howard2017mobilenets,Sandler_2018_CVPR} for super-resolution and deblurring.
    In \cite{Ahn_2018_ECCV}, efficient residual block design was presented while \cite{Kupyn_2019_ICCV} applied the depthwise separable convolution.
    Recently, \cite{Li_2019_ICCV} proposed to linearly decompose the convolutional filters and learn the basis of filters with optional basis sharing between layers.
    These methods effectively reduces the model computation while preserving the restoration accuracy in a similar level.
    
    
    PIRM 2018~\cite{Ignatov_2018_ECCV_Workshops} and AIM 2019~\cite{Zhang_2019_ICCV_Workshops} organized challenges for image super-resolution and image enhancement on smartphones and for constrained super-resolution, respectively. We refer the reader to~\cite{Ignatov_2018_ECCV_Workshops_AI,Ignatov_2019_ICCV_Workshops_AI} for an overview and benchmark of Android smartphones and their capability to run deep learning models.


\section{The Challenge}
\label{sec:challenge}

    We hosted the NTIRE 2020 Image and Video Deblurring Challenge in order to promote developing the state-of-the-art algorithms for image and video deblurring.
    Following the NTIRE 2019 Challenge on Video Deblurring and Super-Resolution, we use the REDS dataset~\cite{Nah_2019_CVPR_Workshops_REDS} to measure the performance of the results. 
    
    \subsection{Tracks}
    \label{sec:tracks}
        In this challenge, we divide the competition into 3 tracks: (1) Image Deblurring (2) Image Deblurring on Mobile Devices (3) Video deblurring.
        
        \noindent \textbf{Track 1: Image Deblurring} aims to develop single-image deblurring methods without limiting the computational resources.
        
        \noindent \textbf{Track 2: Image Deblurring on Mobile Devices} goes beyond simply developing well-performing single-image deblurring methods.
        To encourage steps towards more practical solutions, the running time is measured as well as the restoration accuracy.
        The challenge participants are requested to submit TensorFlow Lite models so that the running speed is measured by the organizers.
        
        We use Google Pixel 4 (Android 10) as the platform to deploy the deblurring models.
        The processor is Qualcomm snapdragon 855 which supports GPU and DSP acceleration.
        32-bit floating point models typically choose to run on GPU while the 8-bit quantized models further accelerate on DSP via Android NNAPI.
        
        \noindent \textbf{Track 3: Video Deblurring} targets developing video deblurring methods that exploit temporal relation between the video frames.
        The winner of the NTIRE 2019 challenge, EDVR~\cite{Wang_2019_CVPR_Workshops} learns to align the input feature and performs deblurring afterwards.
        In this challenge, there were several attempts to use additional modules on top of EDVR.
    
    \subsection{Evaluation}
    \label{sec:evaluation}
        The competition consists of development and testing phases.
        During the development phase, the registered participants train their method and could get feedback from online server to validate the solution format. 
        Meanwhile, they can get local feedback directly from the validation data.
        In the testing phase, the participants submit their results as well as the source code, trained model, and the fact sheets describing their solution. 
        The reproducibility is checked by the organizers.
        
        The results are basically evaluated by the conventional accuracy metrics: PSNR and SSIM~\cite{wang2004image}.
        For track 2, we use a score term that favors both fast and accurate methods.
        Compared with our baseline model, we add a relative fps score to the PSNR of the restoration results.
        Finally, our score functions is
        
        \begin{equation}
            \mathrm{score} = \mathrm{PSNR} + \frac{1}{2} \log_{2}{\frac{\mathrm{FPS}}{\mathrm{FPS}_{baseline}}},
            \label{eq_score_track2}
        \end{equation}
        
        where our baseline is a simple residual network with 4 blocks and 64 channels~\cite{Lim_2017_CVPR_Workshops}.
        $\mathrm{FPS}_{baseline}$ is 8.23.
        The methods faster than the baseline gets higher score than its PSNR while the slower are penalized.
        The gains from acceleration is evaluated in log scale to prevent extremely fast methods without meaningful processing
        Also, we set the maximum scorable fps to be 45.
        However, no submission score was plateaued by the fps limit.
        
\section{Challenge Results}

    Each challenge track had 163, 135, and 102 registered participants. 
    In the final testing phase, total 9, 5, 8 results were submitted.
    The teams submitted the deblurred frames as well as the source code and the trained models.
    
    Table~\ref{table:track1}, \ref{table:track2}, \ref{table:track3} each summarizes the result of the corresponding competition track.
    All the proposed methods on desktop use deep neural networks with GPU acceleration. 
    Mobile accelerators are shown in Table~\ref{table:track2}.
    
    \noindent \textbf{Baseline methods} We present the baseline method results to compare with the participants' methods.
    For track 1, we present the result of Nah~\etal~\cite{Nah_2017_CVPR} that is trained with the REDS dataset. L1 loss is used to train the model for 200 epochs with batch size 16. The learning rate was set to $10^{-4}$ and halved at 100th, 120th, and 140th epoch.
    For track 2, we present a EDSR~\cite{Lim_2017_CVPR_Workshops}-like architecture without upscaling module. 4 ResBlocks are used with 64 channels.
    For track 3, we compare the results with EDVR~\cite{Wang_2019_CVPR_Workshops}, the winner of NTIRE 2019 Video Deblurring Challenge.
    
    \noindent \textbf{Architectures and Main ideas} 
    There were several attempts to use multi-scale information in different perspectives.
    UniA team used atrous convolution~\cite{yu2015multi} and uses multi-scale aggregation for video deblurring
    MTKur, Wangwang, CVML, VIDAR, Reboot, Duke Data Science chose encoder-decoder or U-net style architecture while Vermilion used SRN~\cite{Tao_2018_CVPR}.
    In contrast, IPCV\_IITM, CET\_CVLab, Neuro\_avengers adopted DMPHN~\cite{Zhang_2019_CVPR} without scaling.
    On the other hand, OIerM used fractal architecture to fuse multi-depth information.
    
    For video deblurring, EMI\_VR, IMCL-PROMOTION modified EDVR~\cite{Wang_2019_CVPR_Workshops}.
    UIUC-IFP modified their previous WDVR~\cite{Fan_2019_CVPR_Workshops} by frame concatenation.
    
    Several teams used loss other than L1 or L2 loss: adversarial loss (CET Deblurring Team, SG), WAE-MMD loss (CVML), perceptual loss(IMCL-PROMOTION, SG).
    
    \noindent \textbf{Restoration fidelity} UniA Team, MTKur are the winners of NTIRE 2020 challenge in track 1 and 2.
    In track 3, the submitted results did not improve from the NTIRE 2019 winner, EDVR.

\begin{table*}[ht]
    \centering
    \footnotesize
    {
    \begin{tabular}{l l | c c c c}
        \multicolumn{2}{c|}{} & \multicolumn{4}{c}{Track 1: Image Deblurring} \\
        Team & Method & PSNR & SSIM & FPS & GPU\\
        \hline
        UniA Team & Wide Atrous Block (ensemble~$\times4$)& 34.44 & 0.9412 & 1.43 & Tesla V100\\    
        OIerM & Attentive Fractal Network & 34.20 & 0.9392 & 1.16 & RTX 2080 Ti\\
        MTKur & DRU-prelu~(ensemble~$\times3$) & 33.35 & 0.9283 & 0.83 & RTX 2080 Ti\\
        Wangwang & Two-stage EdgeDeblurNet & 33.07 & 0.9242 & 0.46 & GTX 1080 Ti\\
        IPCV\_IITM & DMPHN + Region-Adaptive Network~(ensemble~$\times8$) & 33.03 & 0.9242 & 0.56 & RTX 2080 Ti\\
        Vermilion & Simplified SRN (ensemble~$\times3$) & 30.04 & 0.8616 & 0.36 & GTX 1080 (eGPU)\\
        CET\_CVLAB & Stack-DMPHN & 29.78 & 0.8629 & 0.50 & Quadro K6000\\
        CVML & Wasserstein Autoencoder & 28.10 & 0.8097 & 33.3 & RTX 2080 SUPER\\
        CET Deblurring Team & DoubleGAN & 26.58 & 0.7492 & 2.04 & GTX 1080\\
        \hdashline[1pt/2pt]
        \textit{baseline} & Multi-scale CNN~\cite{Nah_2017_CVPR}\footnotemark[1] & 32.90 & 0.9207 & 1.12 & RTX 2080 Ti\\
        \textit{baseline} & no processing & 26.13 & 0.7749 & - & -\\
    \end{tabular}
    \caption{Single image deblurring results on the REDS test data.}
    \label{table:track1}
    \vspace{2mm}
    }

    \centering
    \footnotesize
    \begin{tabular}{l l | c c c c c}
        \multicolumn{2}{c|}{} & \multicolumn{5}{c}{Track 2: Image Deblurring~(Mobile)}\\
        Team & Method & PSNR & SSIM & FPS\footnotemark[2] & Final score & Accelerator\\
        \hline
        \multirow{2}{*}{MTKur}
            & DRU-relu-compressed & 32.07 & 0.9024 & 17.6 & 32.62 & NNAPI\\
            & DRU-prelu & 32.95 & 0.9239 & 5.1 & 32.60 & GPU\\
        VIDAR & Transformed fusion U-Net & 30.20 & 0.8735 & 6.0 & 29.97 & GPU\\
        Reboot & Light-weight Attention Network & 31.38 & 0.8960 & 1.0 & 29.87 & GPU\\
        OIerM & Attentive Fractal Network & 28.33 & 0.8079 & 0.9 & 26.71 & CPU\\
        
        \hdashline[1pt/2pt]
        \textit{baseline} & 4 ResBlocks & 28.46 & 0.8218 & 8.23 & 28.46 & GPU\\
        \textit{baseline} & no processing & 26.13 & 0.7749 & - & - & -\\
    \end{tabular}
    \caption{Single image deblurring results on the REDS test data from Google Pixel 4.}
    \label{table:track2}
    \vspace{2mm}

    \centering
    \footnotesize
    \begin{tabular}{l l | c c c c}
        \multicolumn{2}{c|}{} & \multicolumn{4}{c}{Track 3: Video Deblurring} \\
        Team & Method & PSNR & SSIM & FPS & GPU\\
        \hline
        EMI\_VR & PAFU (ensemble $\times4$) & 36.93 & 0.9649 & 0.14 & Tesla V100\\
        UIUC-IFP & WDVR+ (ensemble $\times8$)& 35.58 & 0.9504 & 0.07 & GTX 1080 Ti \\
        IMCL-PROMOTION & PROMOTION & 35.42 & 0.9519 & 0.94 & GTX 1080 Ti\\
        UniA Team & Dual-Stage Multi-Level Feature Aggregation & 34.68 & 0.9442 & 0.18 & Tesla V100\\
        \multirow{2}{*}{Neuro\_avengers} 
            & DMPHN + GridNet & 31.85 & 0.9074 & 0.40 & Titan X\\
            & DMPHN           & 31.43 & 0.8949 & 0.97 & Titan X\\
        SG & Multi-Loss Optimization & 29.44 & 0.8526 & 0.48 & GTX Titan Black\\
        Duke Data Science & Encoder-Decoder & 26.88 & 0.8051 & 0.24 & Tesla V100\\    
        
        \hdashline[1pt/2pt]
        \textit{NTIRE 2019 winner} & EDVR~\cite{Wang_2019_CVPR_Workshops} & 36.94\footnotemark[3] & 0.9656 & 0.26 & Titan Xp\\
        \textit{baseline} & no processing & 26.13 & 0.7749 & - & -\\
    \end{tabular}
    \caption{Video deblurring results on the REDS test data.}
    \label{table:track3}
    \vspace{-3mm}
\end{table*}

\section{Challenge Methods and Teams}
\footnotetext[1]{trained on REDS dataset without adversarial loss.}
\footnotetext[2]{$256\times256$ resolution is used as processing on mobile devices is slow.}
\footnotetext[3]{In NTIRE 2020, boundaries are included in the evaluation.}

We describe the submitted solution details in this section. 

\subsection{MTKur team - Track 1, 2}
\label{sec:MTKur}

\begin{figure}[h]
        \centering
        \includegraphics[width=\linewidth]{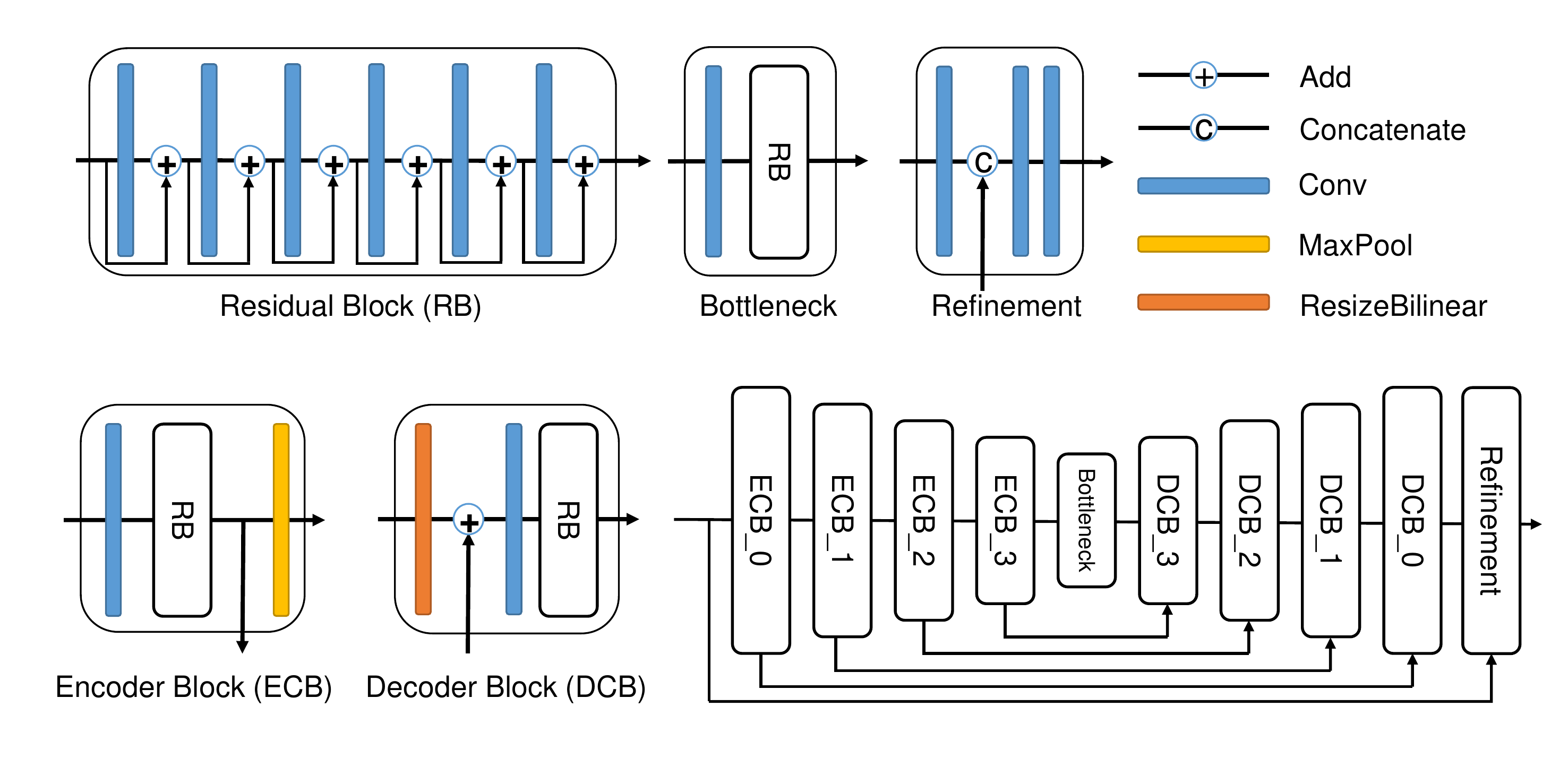}
        \caption{MTKur team: Dense Residual U-Net}
        \label{fig:MTKur}
        \vspace{-3mm}
    \end{figure}

    MTKur team is the winner of Track 2. Following the guidelines in \cite{CVPRW20_DeployOnMobile}, MTKur team develops Dense Residual U-Net (DRU) architecture by applying several modifications to U-Net~\cite{ronneberger2015u}.
    First, they replace concatenation operation in all skip connections with addition operation except the last global connection.
    Second, they replace the single convolution operation with a dense residual block to improve deblurring quality.
    Third, considering the mobile deployment, TransposeConv operations are replaced with ResizeBilinear operations as they have poor latency on Pixel 4.
    The overall DRU architecture is shown in Figure~\ref{fig:MTKur}.
    
    Based on DRU, MTKur proposes 2 variations: DRU-prelu and DRU-relu-compressed depending on the purpose.
    DRU-prelu targets on better restoration quality by using PReLU activation~\cite{He_2015_ICCV} after all convolutional layers except the convolutional layers before ResizeBilinear and the last convolutional layer.
    %
    In contrast, DRU-relu-compressed uses ReLU activation aiming higher throughput of the model. Also, current Tensorflow Lite kernel generates erroneous outputs of quantized PReLU layer.
    
    To DRU-relu-compressed, a series of network compression techniques provided in MediaTek's NeuroPilot toolkit~\cite{neuropilot} are applied including pruning and quantization after basic training.
    First, iterative pruning scheme~\cite{wang2019architecture} exploited. 
    The model repeats to prune by 5\% MAC reduction criterion and retrains to achieve the original PSNR quality. 
    After the iteration, the pruned model achieves 20\% MAC reduction without PSNR drop in the REDS validation dataset.
    Then, the network is further optimized with quantization-aware training of NeuroPilot toolkit \cite{neuropilot} to minimize PSNR drop.
    
    To obtain better results, geometric self-ensemble(x3)~\cite{Timofte_2016_CVPR,Lim_2017_CVPR_Workshops} is applied to DRU-prelu in Track 1.
    In Track 2, ensemble was not used for better running time.
    
    The proposed models are trained with $256\times256$ sized patches of batch size 16.
    $L_{1}$ loss is applied at learning rate $2\times10^{-4}$, exponential decaying by rate 0.98 with 5000 decay steps.
    The input and target images are normalized to range [0, 1]. 
    The training before compression takes 10 days on a single RTX 2080 Ti. Pruning and quantization takes 6 and 2 days, respectively.
    
\subsection{UniA team - Track 1, 3}
\label{sec:UniA}

\begin{figure}[h]
        \centering
        \includegraphics[width=\linewidth]{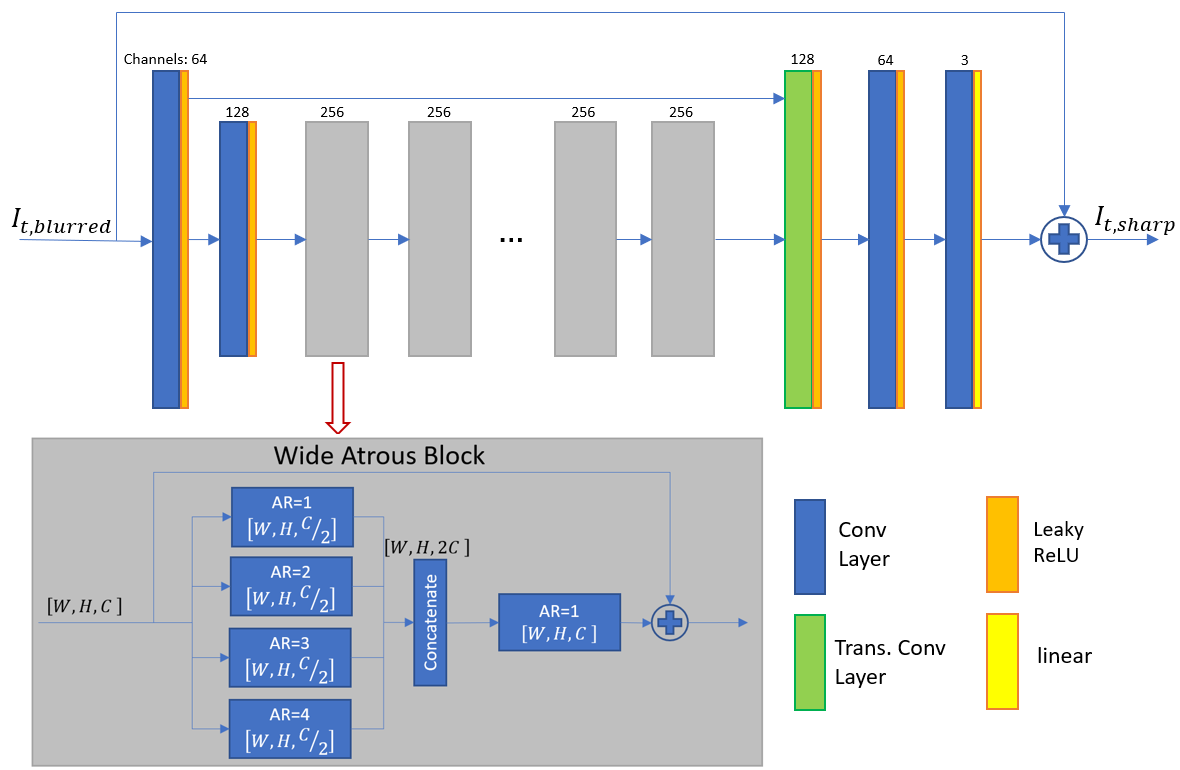}
        \caption{UniA team (Track 1): Atrous Convolutional Network}
        \label{fig:UniA1}
    \end{figure}

    UniA team proposes an image deblurring and a video deblurring method~\cite{BreScher_2020_Dual} for track 1 and 3, respectively.
    %
    In track 1, they propose to use atrous convolution to increase the receptive field instead of downsampling input multiple times to prevent loss of information.
    From the experiments, receptive fields with missing pixels inside were not beneficial.
    Thus, Wide Atrous Block is designed where parallel atrous convolutions with different dilation rates are used.
    The features are concatenated to be used in the next layer.
    The model architecture is shown in \ref{fig:UniA1}.
    To stabilize the training, the convolutional features are scaled by learnable parameters and added by a constant before the activation.
    LeakyReLU activation is used except the last layer which uses a linear activation.
    
    \begin{figure}[h]
        \centering
        \includegraphics[width=\linewidth]{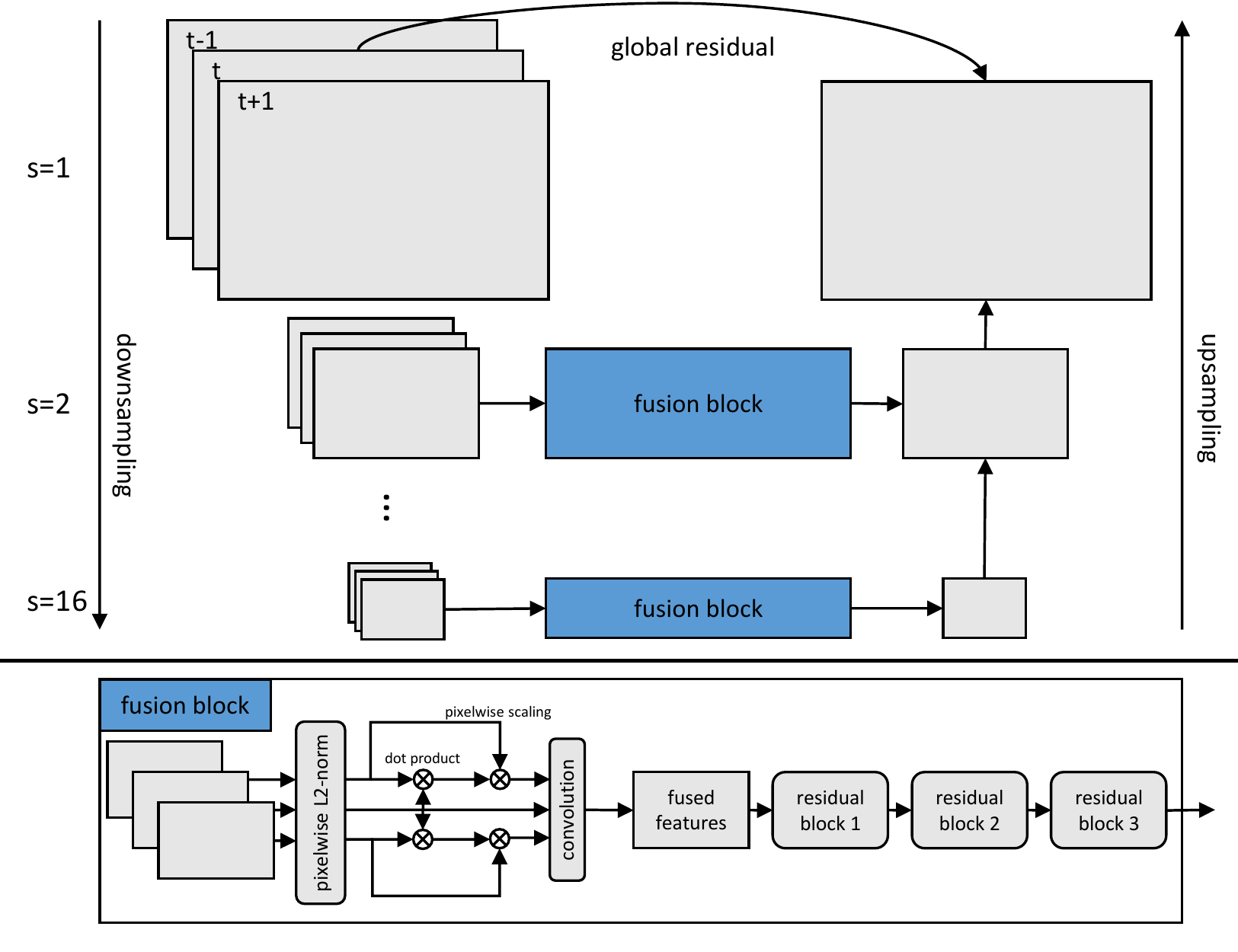}
        \caption{UniA team (Track 3): Dual-Stage Multi-Level Feature Aggregation}
        \label{fig:UniA2}
        \vspace{-5mm}
    \end{figure}
    
    The model is trained with extensive data augmentation including brightness, contrast, hue augmentation.
    Patch size is set to $320\times320$. 
    VGG, edge similarity, adversarial loss were investigated but didn't bring improvements in terms of PSNR an d SSIM.
    Geometric self-ensemble is applied to improve performance~\cite{Timofte_2016_CVPR,Lim_2017_CVPR_Workshops}.
    
    The video deblurring method performs post-processing to the results obtaine from the image deblurring method.
    The features from the target and the neighboring frame are fused in multiple scales.
    From downsampled frames, independent convolutional features are concatenated and fused by residual blocks.
    The coarse features are upsampled and added to the higher resolution features. 
    Figure~\ref{fig:UniA2} shows the video deblurring architecture.

\subsection{OIerM team - Track 1, 2}
\label{sec:OIerM}

\begin{figure}[h]
        \centering
        \includegraphics[width=\linewidth]{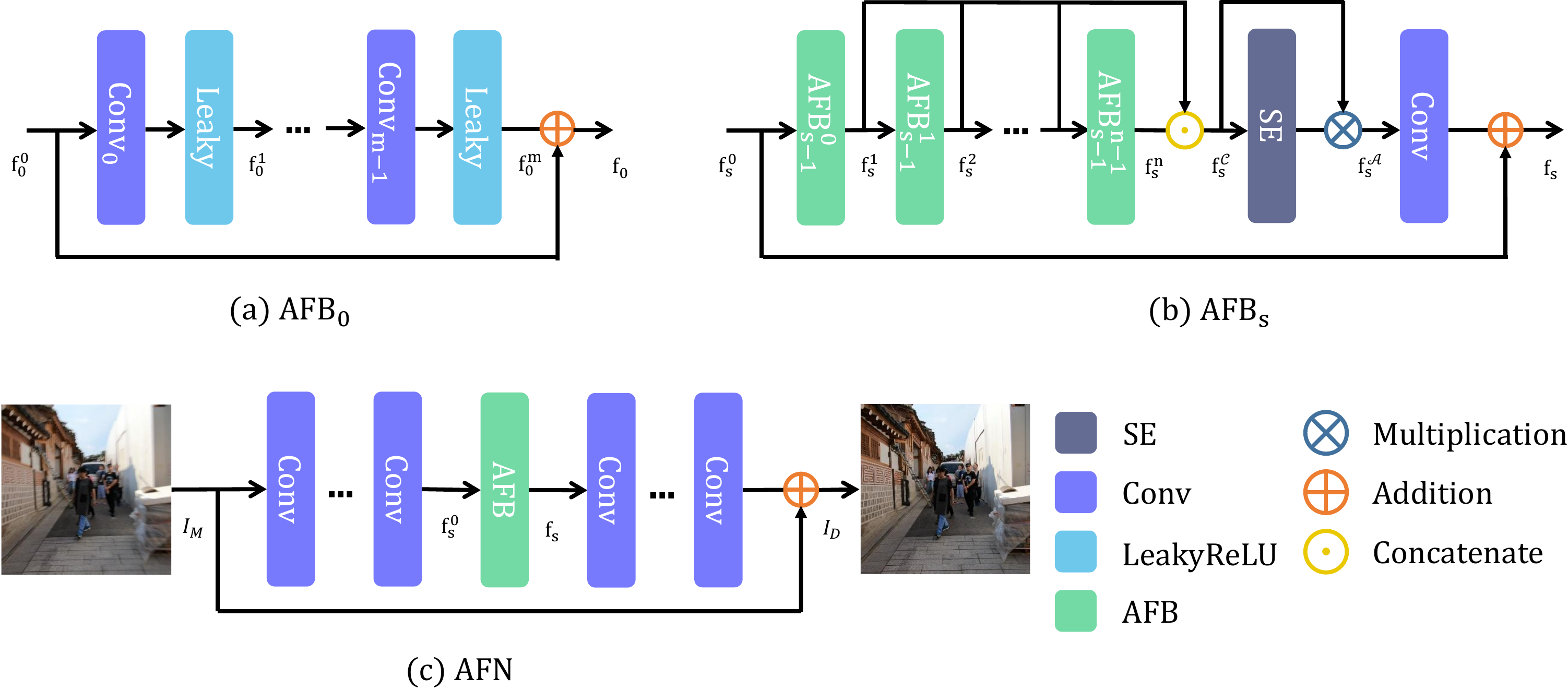}
        \caption{OIerM team: Attentive Fractal Network}
        \label{fig:OIerM}
        \vspace{-3mm}
    \end{figure}

    OIerM team proposes an Attentive Fractal Network~(AFN)~\cite{OIerM_2020_Demoire}. 
    They construct the Attentive Fractal Block via progressive feature fusion and channel-wise attention guidance. 
    Then AFB is stacked in a fractal way inspired by FBL-SS~\cite{yang2019towards} such that a higher-level $AFB_{s}$ is constructed with the lower-level $AFB_{s-1}$, recursively with self-similarity, as shown in Figure~\ref{fig:OIerM}. 
    Shortcuts and residual connections at different scale effectively resolve the vanishing gradients and help the network learn more key features. 
    The progressive fusion of intermediate features let the network handle rich information.

    For track 1 and 2, $n$, $m$, and $s$ in Figure~\ref{fig:OIerM} are set to 4 and 2, respectively.
    The models are trained with batch size 8 and 16 for 50 and 200 epochs.
    The channel size is 128.

\subsection{Wangwang team - Track 1}
\label{sec:Wangwang}

Wangwang team proposes a solution, EdgeDeblurNet, based on the DeblurNet module of DAVANet~\cite{Zhou_2019_CVPR}.
As the edges and smooth area are differently blurred, they employ the edge map as extra available information to guide deblurring.
To better emphasize the main object edges than the blur trajectories, spatial and channel attention gate~\cite{Woo_2018_ECCV} is adopted.
Also, SFT-ResBlock~\cite{Wang_2018_CVPR} is used to fuse the edge information effectively.
Finally, inspired by the traditional iterative optimization methods, a two-stage deblurring method is proposed by cascading the EdgeDeblurNet.

\begin{figure}[h]
    \centering
    \includegraphics[width=\linewidth]{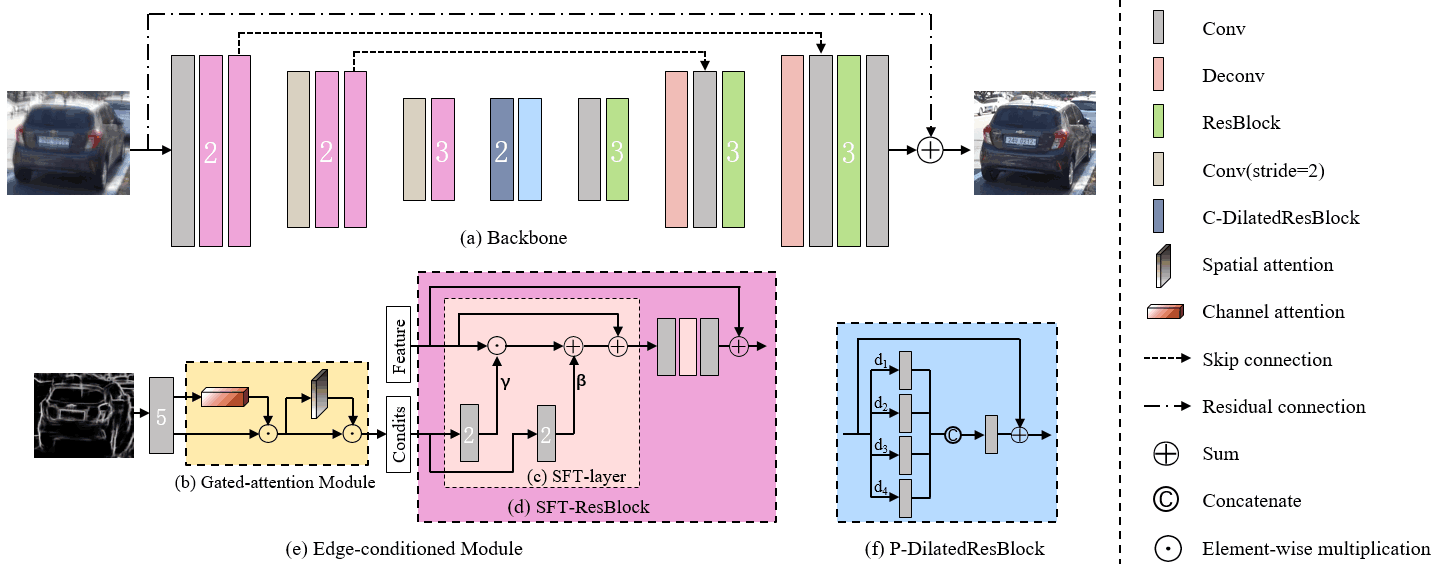}
    \caption{Wangwang team: Two-stage Edge-Conditioned Network}
    \label{fig:Wangwang}
    \vspace{-3mm}
\end{figure}

The second stage model is trained after the first stage parameters are trained and fixed.
The model is trained with $256\times256$ patches of batch size 16 by Adam~\cite{kingma2014adam} optimizer for 400 epochs.
The learning rate is initialized as $10^{-4}$ and decayed at predetermined steps.
The overall architecture is shown in Figure~\ref{fig:Wangwang}.

\subsection{IPCV\_IITM team - Track 1}

\begin{figure}[h]
    \centering
    \includegraphics[width=\linewidth]{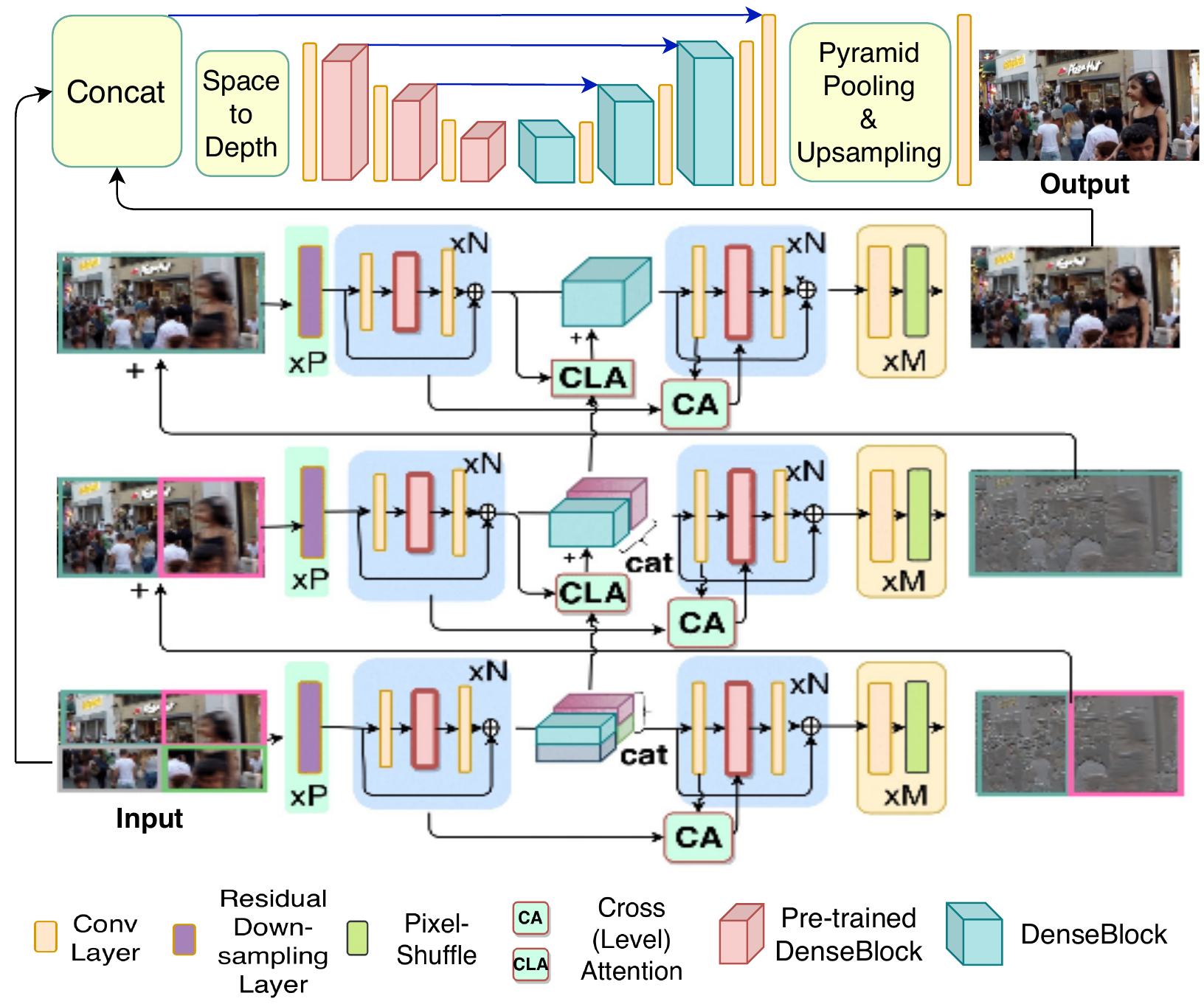}
    \caption{IPCV\_IITM team: Region-Adaptive Patch-hierarchical Network}
    \label{fig:IPCV_IITM}
    \vspace{-3mm}
\end{figure}

IPCV\_IITM team buids a two-stage network based on DMPHN~\cite{Zhang_2019_CVPR} and \cite{suin2020spatiallyattentive}.
The first stage is a 3-level DMPHN with cross-attention modules similar to \cite{suin2020spatiallyattentive} where each pixel can gather global information.
The second stage is a densely connected encoder-decoder structure~\cite{purohit2019region}. 
Before the last layer, The multi-scale context aggregation is performed through pooling and upsampling.
Geometric self-ensemble~\cite{Timofte_2016_CVPR,Lim_2017_CVPR_Workshops} is applied to further improve results.
Figure~\ref{fig:IPCV_IITM} shows the overall architecture.

\subsection{Vermilion team - Track 1}

Vermilion team uses a simplified Scale-Recurrent Networks~\cite{Tao_2018_CVPR} in 4 scale levels. 
To make SRN simpler, the recurrent connections are removed. 
On each level, a series of ResBlocks are used instead of U-Net based structure. 
The final deblurred result is  obtained by an ensemble of 3 results.

The model is trained with $256\times256$ patches of batch size 4.
The learning rate was set to $10^{-4}$.



\begin{figure}[h]
    \centering
    \includegraphics[width=\linewidth]{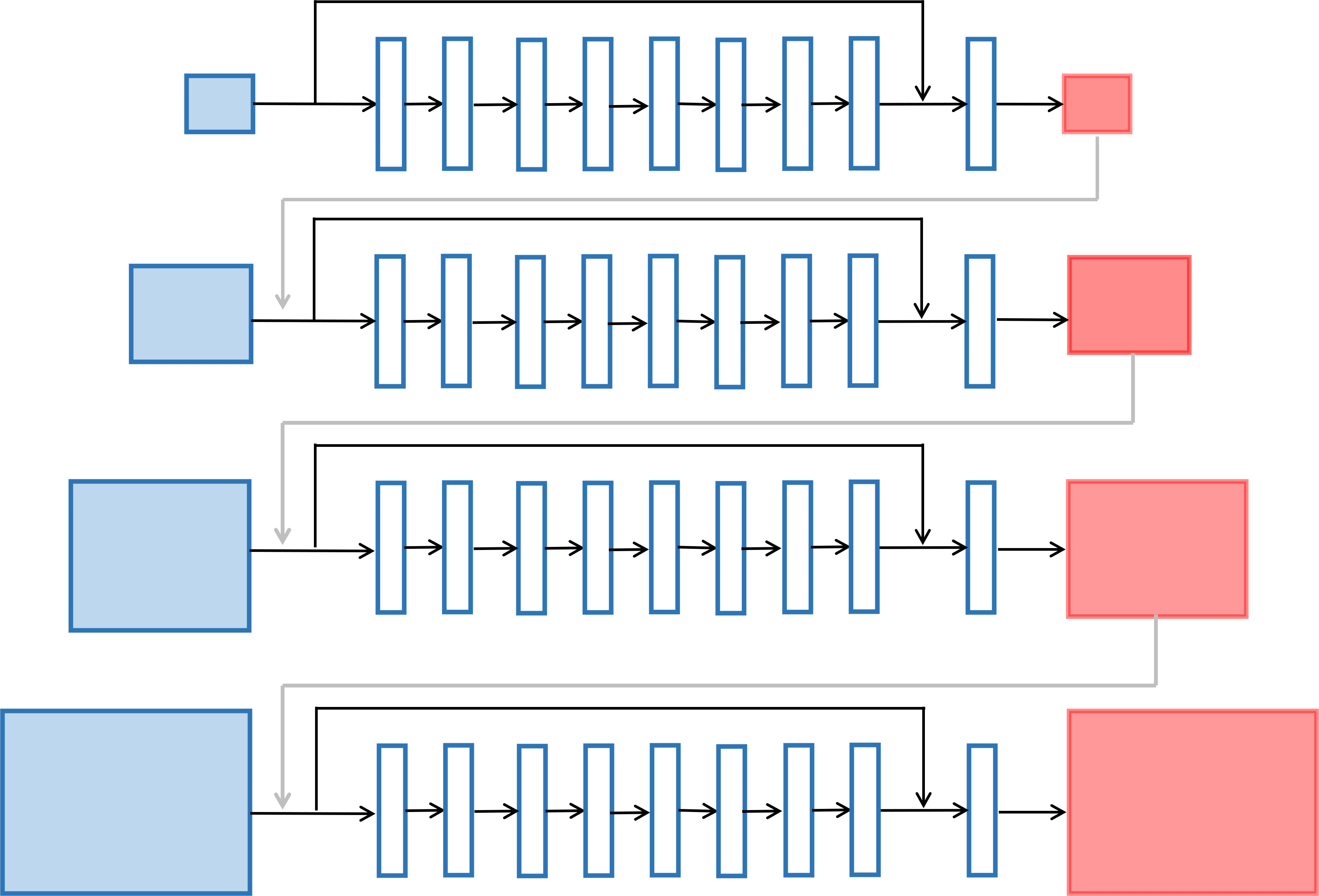}
    \caption{Vermilion team: Simplified SRN architecture}
    \label{fig:Vermilion}
    \vspace{-3mm}
\end{figure}

\subsection{CET\_CVLab team - Track 1}
\label{sec:CET_CVLab}

CET\_CVLab team uses Stack-DMPHN~\cite{Zhang_2019_CVPR} with 5 levels.
The upsampling method is replaced by depth-to-space operation.
The model is trained with MSE loss with Adam optimizer.
The learning rate is initialized with $10^{-4}$ and halved at every 100 epochs.



\subsection{CVML team - Track 1}
\label{sec:CVML}

\begin{figure}[h]
        \centering
        \includegraphics[width=\linewidth]{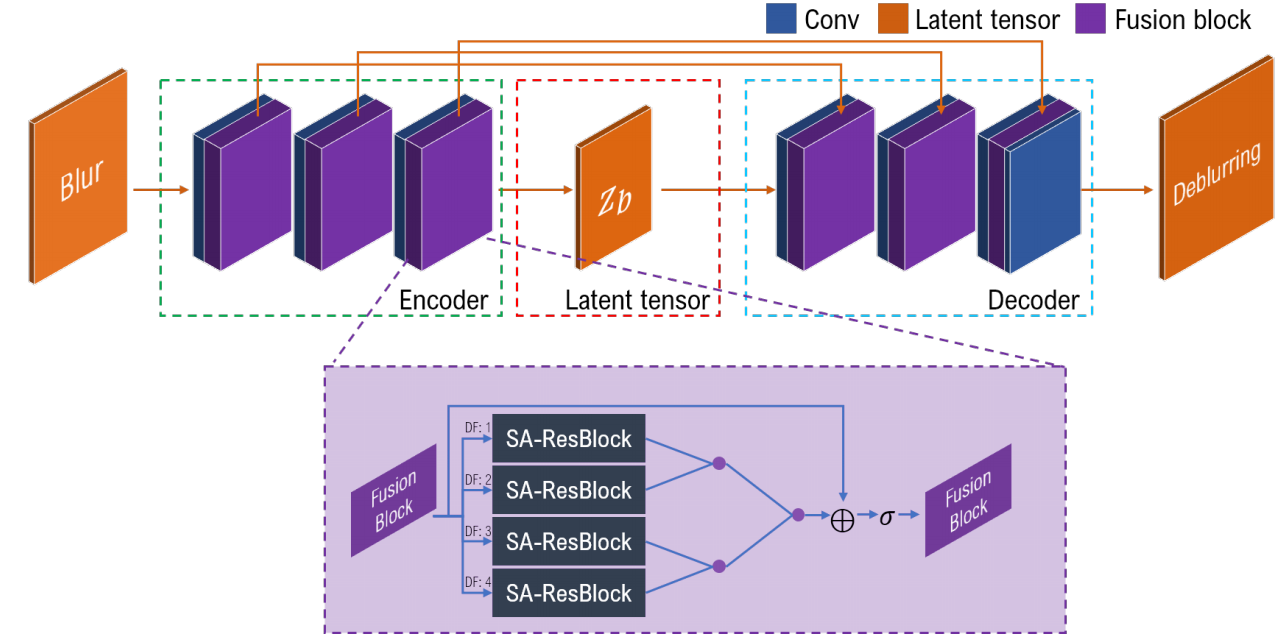}
        \caption{CVML team: Wasserstein Autoencoder}
        \label{fig:CVML}
        \vspace{-3mm}
    \end{figure}

    CVML team uses a Wasserstein autoencoder~\cite{tolstikhin2017wasserstein} for single image deblurring.
    The latent space is represented as spatial tensor instead of 1D vector.
    In addition, an advanced feature based architecture is designed to deliver rich feature information to the latent space.
    There are three tree-structured fusion modules for encoder and decoder, respectively.
    To train the model, WAE-maximum mean discrepancy (WAE-MMD) loss, reconstruction loss, and perceptual loss is applied to the $240\times240$ patches.
    The overall architecture is shown in Figure~\ref{fig:CVML}.
    
    
\subsection{CET Deblurring team - Track 1}

\begin{figure}[h]
        \centering
        \includegraphics[width=\linewidth]{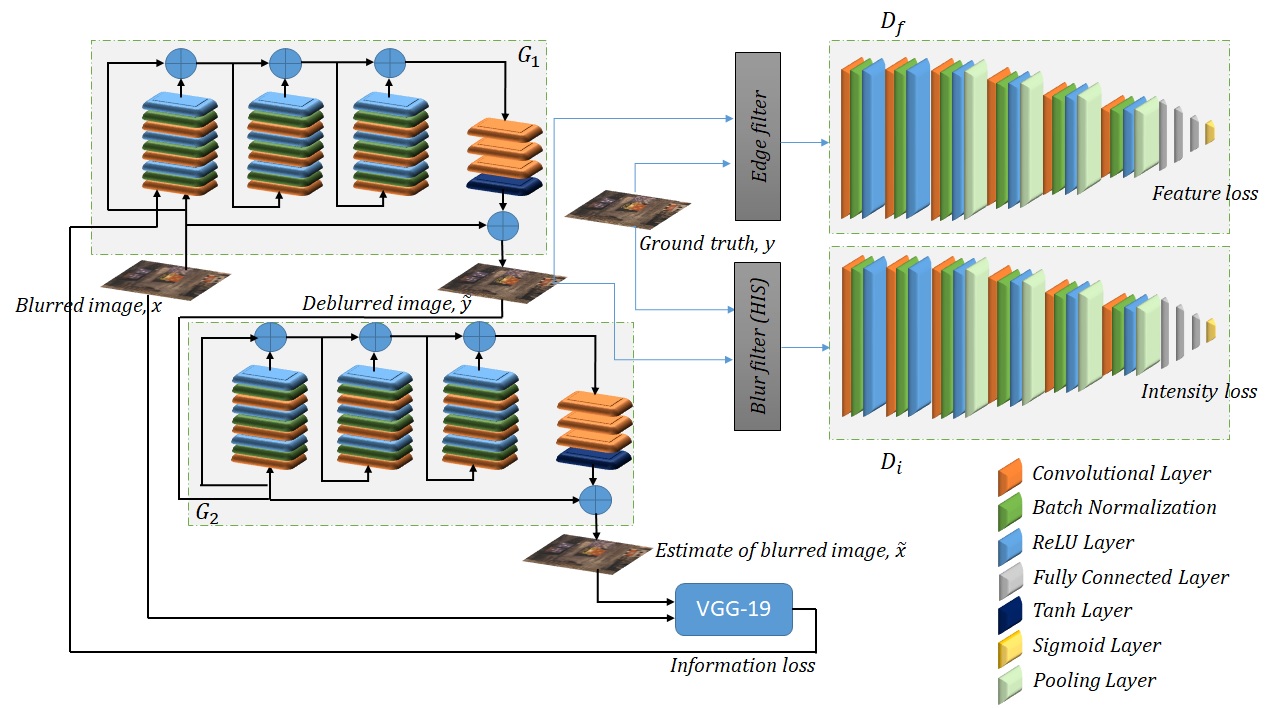}
        \caption{CET\_Deblurring team: Double GAN}
        \label{fig:CET_Deblurring}
        \vspace{-3mm}
    \end{figure}

    CET Deblurring team proposes a double generative adversarial network that consists of two generators and discriminators.
    The generator $G_{1}$ and $G_{2}$ performs image deblurring and the inverse operation.
    As $G_{2}$ is trained with VGG-19~\cite{simonyan2014very} loss to recover blurry image, $G_{1}$ learns to deblur the original and the re-blurred.
    The two discriminators try to classify the output from $G_{1}$ and the ground-truth with different objectives.
    $D_{f}$ focuses on the edges by looking at the gray LoG features while $D_{I}$ refers to the color intensity in HSI format from Gaussian blurred images.
    The overall architecture is shown in Figure~\ref{fig:CET_Deblurring}.

\subsection{VIDAR team - Track 2}

    VIDAR team proposes a Transformed fusion U-Net. 
    In the connections between the corresponding encoder and decoder, the encoder features go through a transform layer to get useful information. 
    Also, the decoder features from different scales are concatenated together. 
    Finally, different dilated convolution layers are used that are similar to atrous pyramid pooling to estimate different blur motion. 
    
    To train the model, L1 loss, SSIM loss, and multi-scale loss are applied.
    The training patch size is $256\times256$ and the batch size is 16.
    Adam optimizer is used with learning rate decaying from $2.2\times10^{-5}$ by power $0.997$.
    Training stops when the training loss does not notably decreases.
    




\subsection{Reboot team - Track 2}

\begin{figure}[h]
        \centering
        \includegraphics[width=\linewidth]{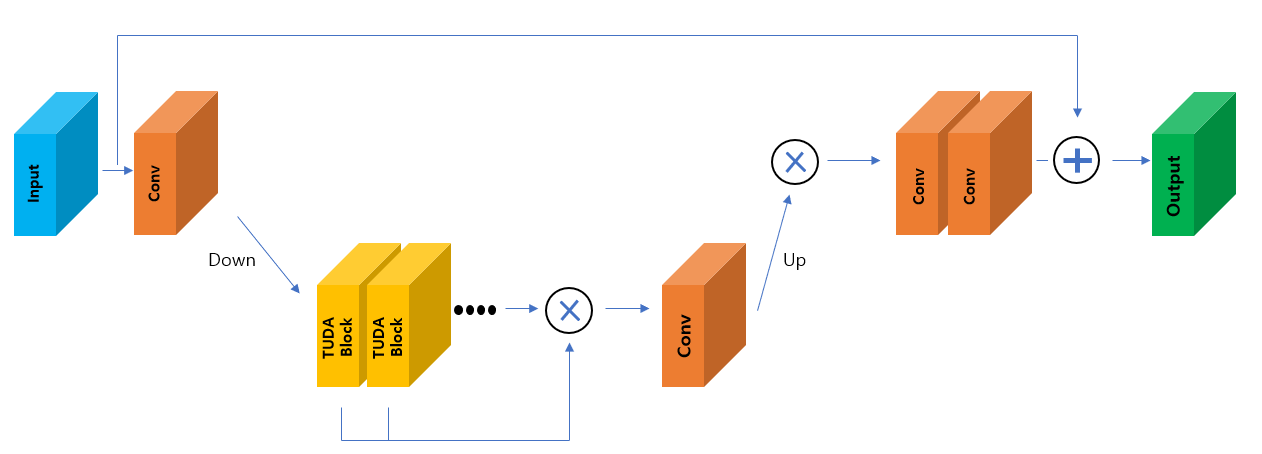}
        \caption{Reboot team: Light-weight Attention Network}
        \label{fig:Reboot}
        \vspace{-3mm}
    \end{figure}

    Reboot team proposes a light-weight attention network from their proposed building blocks.
    TUDA (Total Up-Down Attention) block is the high-level block that contains other blocks: UAB and UDA.
    UAB (U-net based Attention Block) resembles U-net like architecture that replaces convolutions with UDA.
    UDA (Up-Down Attention block) is the basic block that operates like ResBlock but in a downsampled scale for efficiency.
    The overall architecture is shown in Figure~\ref{fig:Reboot}.

    To train the model, $96\times96$ patches are used with batch size 32.
    Adam optimizer is used with learning rate beginning from $10^{-4}$ that is halved every 50 epochs.
    
\subsection{EMI\_VR team - Track 3}

\begin{figure}[h]
        \centering
        \includegraphics[width=\linewidth]{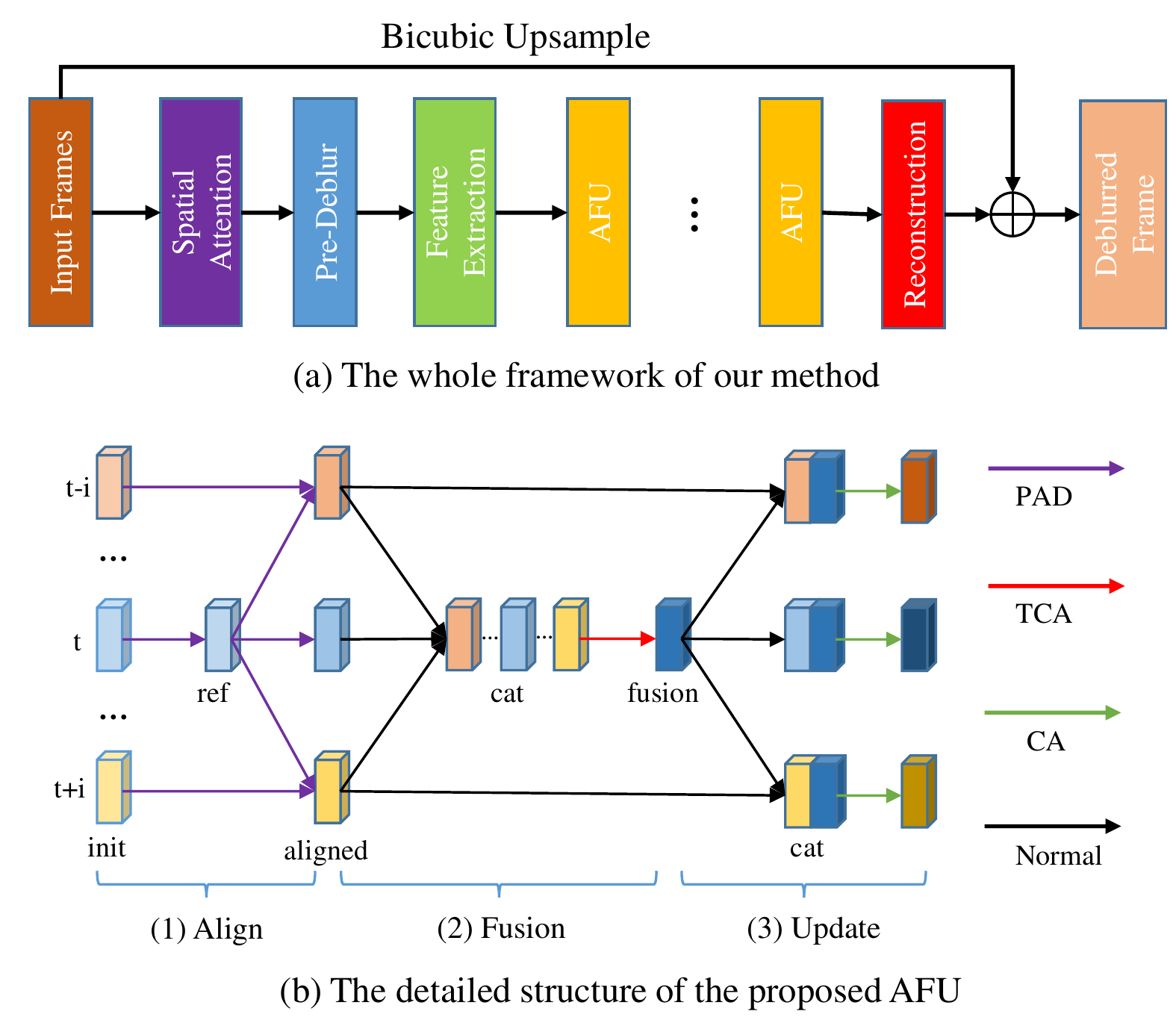}
        \caption{EMI\_VR team: PAFU}
        \label{fig:EMI_VR}
    \end{figure}

    EMI\_VR team proposes a framework, PAFU, that tries to refine the result of 2 serial stack of EDVR~\cite{Wang_2019_CVPR_Workshops}.
    PAFU architecture is similar to EDVR but has different modules.
    First, PAFU uses non-local spatial attention before PreDeblur operation.
    After the PreDeblur process, a series of AFU (Alignment, Fusion, and Update) modules perform progressly on the extracted feature.
    AFU module performs FAD (Feature Align with Deformable convolution) and TCA (Temporal Channel Attention) to fuse aligned features.
    Figure~\ref{fig:EMI_VR} shows the overall architecture of PAFU.
    
    
    
    To train the model, both the training and validation data are used except 4 training videos that are selected by the authors. 
    At test time, geometric self-ensemble~\cite{Timofte_2016_CVPR,Lim_2017_CVPR_Workshops} using flips are used.
    
    
\subsection{UIUC-IFP team - Track 3}

    UIUC-IFP team proposes WDVR+ based on WDVR~\cite{Fan_2019_CVPR_Workshops}.
    In the 2D convolutional WDVR, the input frames were stacked along the channel axis.
    WDVR+ extends WDVR by motion compensation in feature space via template matching.
    Given a patch of the center frame as a template, the most correlated patches in the neighbor frames are searched.
    Such patches are stacked as input for the deblurring model.
    The template matching process is jointly learned with the deep networks.
    At test time, geometric self-ensemble~\cite{Timofte_2016_CVPR,Lim_2017_CVPR_Workshops} is used.

\subsection{IMCL-PROMOTION team - Track 3}

\begin{figure}[h]
        \centering
        \includegraphics[width=\linewidth]{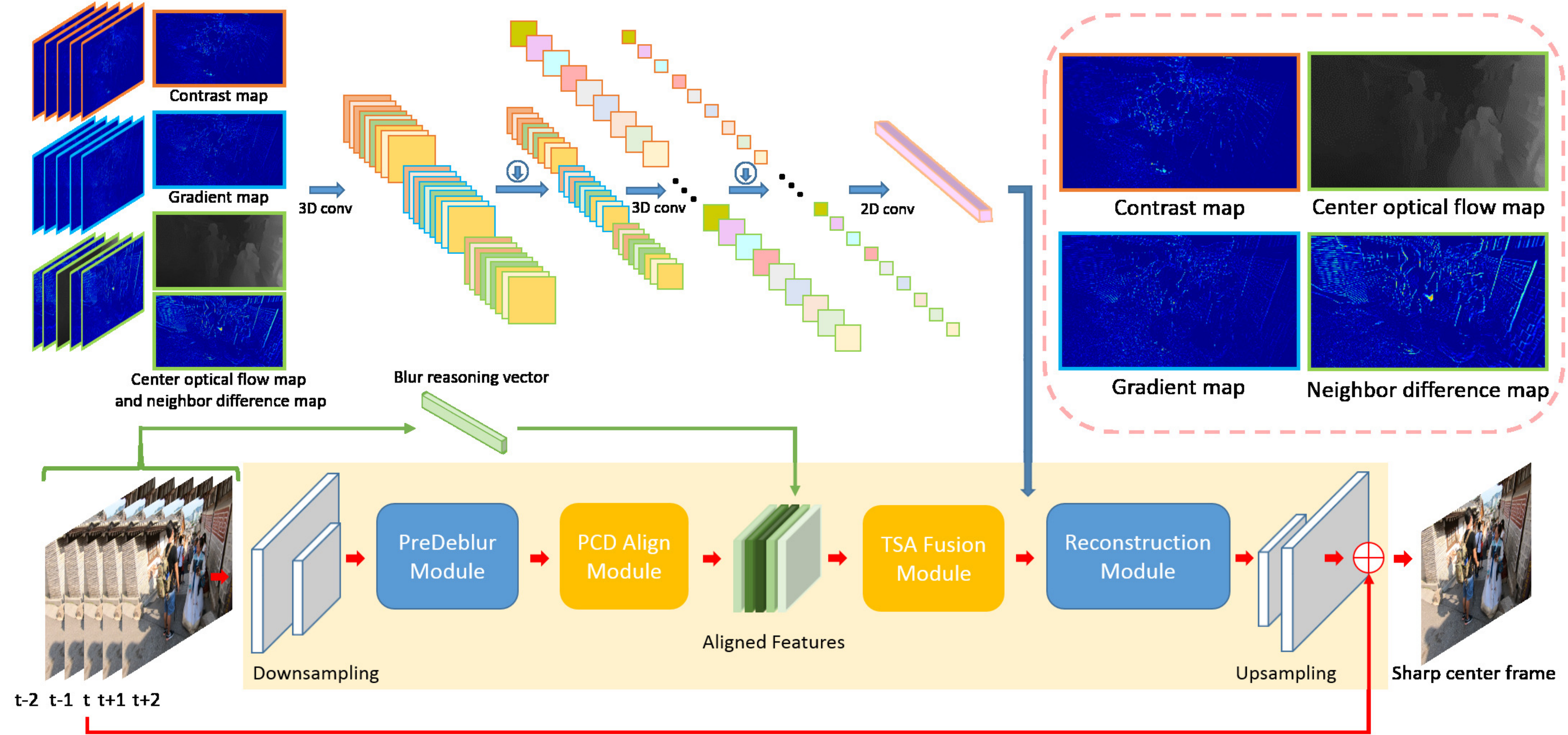}
        \caption{IMCL-PROMOTION team: PROMOTION}
        \label{fig:IMCL}
    \end{figure}

    IMCL-PROMOTION team proposes a framework named PROMOTION~\cite{zhou2020prior} based on the pretrained EDVR~\cite{Wang_2019_CVPR_Workshops} architecture..
    When deblurring models are not properly trained, they may expect the inputs to be always blurry and worsen several sharp inputs by unnecessary operation.
    To let the model distinguish blurry and sharp input, blur reasoning vector is designed to represent the blur degree of each frame.
    Also, to better handle non-uniform nature of motion blur, extra features are extracted.
    From the input frames, image contrast and gradient maps are extracted. 
    To encode the motion information, optical flow of the center frame is calculated by FlowNet 2.0~\cite{Ilg_2017_CVPR}.
    As calculating optial flow for all the frames is heavy, for the neighbor frames, frame difference is used instead.
    Those maps are fed into 3D convolutional networks to encode the heterogeneous information.
    The extracted feature is used at the reconstruction module.
    The overall architecture is shown in Figure~\ref{fig:IMCL}.
    To train the model, Charbonnier loss~\cite{charbonnier1994two} and the perceptual loss~\cite{Zhang_2018_CVPR_unreasonable}.

    
    


\subsection{Neuro\_avengers team}

    Neuro\_avengers team proposes 2 methods by modifying DMPHN~\cite{Zhang_2019_CVPR}.
    In the bottom level, a set of subsequent frames are concatenated while from the higher levels recover the target frame only.
    %
    They propose the second by refining the result of method 1 by cascading GridNet~\cite{Niklaus_2018_CVPR}.
    GridNet takes the 2 neighboring frames warped by optical flow~\cite{Sun_2018_CVPR} as well.
    

\subsection{SG team - Track 3}

\begin{figure}[h]
        \centering
        \includegraphics[width=\linewidth]{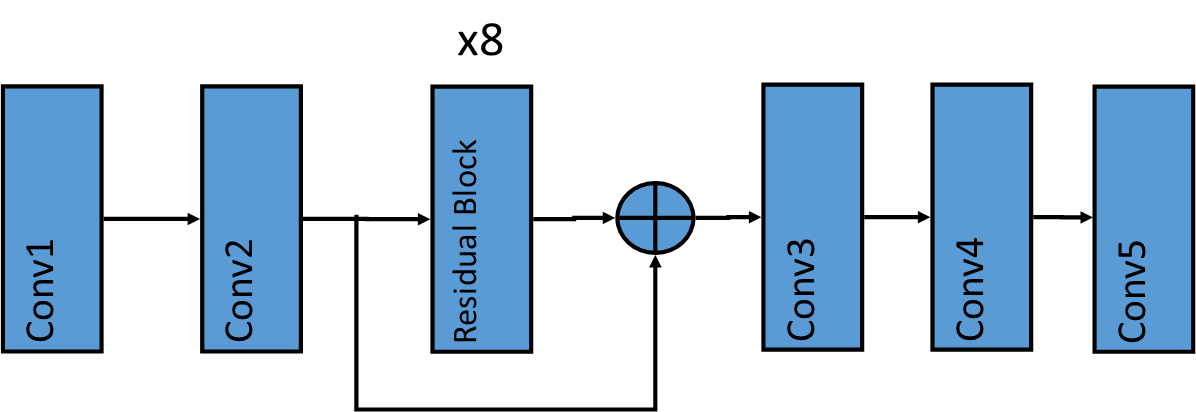}
        \caption{SG team: Multi-Loss Optimization}
        \label{fig:SG1}
    \end{figure}

    SG team trains their model with multiple loss functions: L1, VGG-16, SSIM, and adversarial loss.
    Each convolution is followed by batch normalization~\cite{ioffe2015batch} and PReLU except the first two layers without BN.
    The overall architecture is shown in Figure~\ref{fig:SG1}.
    To handle temporal relation, multiple frames were concatenated as input to the network.
    
\subsection{Duke Data Science team - Track 3}

Duke Data Science team implemented a simple encoder-decoder network based on Su~\etal~\cite{Su_2017_CVPR}. 
The input to the network is a stack of 5 consecutive video frames separated in the color channel. 
There are 3 independent encoder-decoder modules that handle a single color channel. 


\section*{Acknowledgments}

We thank the NTIRE 2020 sponsors: HUAWEI Technologies Co. Ltd., OPPO Mobile Corp., Ltd., Voyage81, MediaTek Inc., DisneyResearch$\mid$Studios, and ETH Zurich (Computer Vision Lab).

\appendix
\section{Teams and affiliations}
\label{sec:appendix}
\subsection*{NTIRE 2020 team}
\noindent\textit{\textbf{Title: }} NTIRE 2020 Challenge on Image and Video Deblurring\\
\noindent\textit{\textbf{Members: }} \textit{Seungjun Nah$^1$ (seungjun.nah@gmail.com)}, Sanghyun Son$^1$, Radu Timofte$^2$,  Kyoung Mu Lee$^1$\\
\noindent\textit{\textbf{Affiliations: }}\\
$^1$ Department of ECE, ASRI, SNU, Korea\\
$^2$ Computer Vision Lab, ETH Zurich, Switzerland\\

\subsection*{MTKur}
\noindent\textit{\textbf{Title 1: }} Dense Residual U-Net for Single Image Deblurring\\
\noindent\textit{\textbf{Title 2: }} Toward Efficient Dense Residual U-Net for Single Image Deblurring\\
\noindent\textit{\textbf{Members: }} \textit{Cheng-Ming Chiang (jimmy.chiang@mediatek.com)}, Yu Tseng, Yu-Syuan Xu, Yi-Min Tsai\\
\noindent\textit{\textbf{Affiliations: }}\\
MediaTek Inc.\\

\subsection*{UniA Team}
\noindent\textit{\textbf{Title 1: }} Atrous Convolutional Block for Image Deblurring\\
\noindent\textit{\textbf{Title 2: }} Dual-Stage Multi-Level Feature Aggregation for Video Deblurring\\
\noindent\textit{\textbf{Members: }} \textit{Stephan Brehm (stephan.brehm@informatik.uni-augsburg.de)}, Sebastian Scherer\\
\noindent\textit{\textbf{Affiliations: }}\\
University of Augsburg, Chair for Multimedia Computing and Computer Vision Lab, Germany\\

\subsection*{OIerM}
\noindent\textit{\textbf{Title: }} Attentive Fractal Network\\
\noindent\textit{\textbf{Members: }} \textit{Dejia Xu$^1$ (dejia@pku.edu.cn)}, Yihao Chu$^2$, Qingyan Sun$^3$\\
\noindent\textit{\textbf{Affiliations: }}\\
$^1$ Peking University, China\\
$^2$ Beijing University of Posts and Telecommunications, China\\
$^3$ Beijing Jiaotong University, China\\


\subsection*{Wangwang}
\noindent\textit{\textbf{Title: }} Two-stage Edge-Conditioned Network for Deep Image Deblurring\\
\noindent\textit{\textbf{Members: }} \textit{Jiaqin Jiang (jiangjiaqin@whu.edu.cn), Lunhao Duan, Jian Yao}\\
\noindent\textit{\textbf{Affiliations: }}\\
Wuhan University, China\\

\subsection*{IPCV\_IITM}
\noindent\textit{\textbf{Title: }} Region-Adaptive Patch-hierarchical Network for Single Image Deblurring\\
\noindent\textit{\textbf{Members: }} \textit{Kuldeep Purohit (kuldeeppurohit3@gmail.com)}, Maitreya Suin, A.N. Rajagopalan\\
\noindent\textit{\textbf{Affiliations: }}\\
Indian Institute of Technology Madras, India\\

\subsection*{Vermilion}
\noindent\textit{\textbf{Title: }} Simplified SRN\\
\noindent\textit{\textbf{Members: }} \textit{Yuichi Ito (wataridori2010@gmail.com)}\\
\noindent\textit{\textbf{Affiliations: }}\\
Vermilion\\

\subsection*{CET\_CVLab}
\noindent\textit{\textbf{Title: }} V-Stacked Deep CNN for Single Image Deblurring\\
\noindent\textit{\textbf{Members: }} \textit{Hrishikesh P S (hrishikeshps@cet.ac.in)}, Densen Puthussery, Akhil K A, Jiji C V\\
\noindent\textit{\textbf{Affiliations: }}\\
College of Engineering Trivandrum\\

\subsection*{CVML}
\noindent\textit{\textbf{Title: }} Image Deblurring using Wasserstein Autoencoder\\
\noindent\textit{\textbf{Members: }} \textit{Guisik Kim (specialre@naver.com)}\\
\noindent\textit{\textbf{Affiliations: }}\\
CVML, Chung-Ang University, Korea\\

\subsection*{CET Deblurring Team}
\noindent\textit{\textbf{Title: }} Blind Image Deblurring using Double Generative Adversarial Network\\
\noindent\textit{\textbf{Members: }} \textit{Deepa P L (deepa.pl@mbcet.ac.in)}, Jiji C V\\
\noindent\textit{\textbf{Affiliations: }}\\
APJ Abdul Kalam Technological University, India\\


\subsection*{VIDAR}
\noindent\textit{\textbf{Title: }} Transformed fusion U-Net\\
\noindent\textit{\textbf{Members: }} \textit{Zhiwei Xiong (zwxiong@ustc.edu.cn)}, Jie Huang, Dong Liu\\
\noindent\textit{\textbf{Affiliations: }}\\
University of Science and Technology of China, China\\

\subsection*{Reboot}
\noindent\textit{\textbf{Title: }} Light-weight Attention Network for Image Deblurring on Smartphone\\
\noindent\textit{\textbf{Members: }} \textit{Sangmin Kim (ksmh1652@gmail.com)}, Hyungjoon Nam, Jisu Kim, Jechang Jeong\\
\noindent\textit{\textbf{Affiliations: }}\\
Image Communication Signal Processing Laboratory, Hanyang University, Korea\\


\subsection*{EMI\_VR}
\noindent\textit{\textbf{Title: }} Progressive Alignment, Fusion and Update for Video Restoration\\
\noindent\textit{\textbf{Members: }} \textit{Shihua Huang (shihuahuang95@gmail.com)}\\
\noindent\textit{\textbf{Affiliations: }}\\
Southern University of Science and Technology, China\\

\subsection*{UIUC-IFP}
\noindent\textit{\textbf{Title: }} WDVR+: Motion Compensation via Feature Template Matching\\
\noindent\textit{\textbf{Members: }} \textit{Yuchen Fan (yc0624@gmail.com)}, Jiahui Yu, Haichao Yu, Thomas S. Huang\\
\noindent\textit{\textbf{Affiliations: }}\\
University of Illinois at Urbana-Champaign\\

\subsection*{IMCL-PROMOTION}
\noindent\textit{\textbf{Title: }} Prior-enlightened and Motion-robust Video Deblurring\\
\noindent\textit{\textbf{Members: }} \textit{Ya Zhou (zhouya@mail.ustc.edu.cn)}, Xin Li, Sen Liu, Zhibo Chen\\
\noindent\textit{\textbf{Affiliations: }}\\
CAS Key Laboratory of Technology in Geo-Spatial Information Processing and Application System, University of Science and Technology of China\\

\subsection*{Neuro\_avengers}
\noindent\textit{\textbf{Title: }} Deep Multi-patch Hierarchical Network for Video Deblurring\\
\noindent\textit{\textbf{Members: }} \textit{Saikat Dutta$^1$ (saikat.dutta779@gmail.com)}, Sourya Dipta Das\\
\noindent\textit{\textbf{Affiliations: }}\\
$^1$ IIT Madra\\
$^2$ Jadavpur University\\

\subsection*{SG}
\noindent\textit{\textbf{Title: }} Multi-Loss Optimization for Video Deblurring\\
\noindent\textit{\textbf{Members: }} \textit{Shivam Garg (shivgarg@live.com)}\\
\noindent\textit{\textbf{Affiliations: }}\\
University of Texas at Austin, USA\\

\subsection*{Duke Data Science}
\noindent\textit{\textbf{Title: }} Encoder-Decoder\\
\noindent\textit{\textbf{Members: }} \textit{Daniel Sprague (dys9@duke.edu)}, Bhrij Patel, Thomas Huck\\
\noindent\textit{\textbf{Affiliations: }}\\
Duke University Computer Science Department\\

{\small
\bibliographystyle{ieee_fullname}

}

\end{document}